\documentclass{article} 
\usepackage{iclr2027_conference,times}

\usepackage{amsmath,amsfonts,bm}

\usepackage{bbm}
\let\oldmathbb\mathbb
\renewcommand{\mathbb}[1]{%
  \ifx1#1\mathbbm{1}%
  \else\oldmathbb{#1}%
  \fi
}

\def\eqref#1{equation~\ref{#1}}

\def\1{\bm{1}}

\DeclareMathAlphabet{\mathsfit}{\encodingdefault}{\sfdefault}{m}{sl}
\SetMathAlphabet{\mathsfit}{bold}{\encodingdefault}{\sfdefault}{bx}{n}

\DeclareMathOperator*{\argmax}{arg\,max}

\usepackage{hyperref}
\usepackage{url}
\usepackage{amsmath}
\usepackage{amssymb}
\usepackage{booktabs}
\usepackage{array}
\usepackage{xcolor}
\usepackage{adjustbox}

\title{Rethinking Circuit Evaluation: Do Circuits Explain Model Errors?}

\author{
Li Zhang$^{*}$ \\
University of Toronto \\
\texttt{zli@cs.toronto.edu}
\And
Chuqin Geng$^{*}$ \\
University of Toronto \\
McGill University \\
\texttt{chuqin.geng@mail.mcgill.ca}
\And
Mark Zhang \\
University of Toronto
\AND
Chen Yang \\
Tsinghua University
\And
Luke Zhang \\
University of Toronto
\And
Haolin Ye \\
McGill University
\And
Xujie Si \\
University of Toronto \\
\texttt{six@cs.toronto.edu}
}

\iclrfinalcopy

\begin{document}

\maketitle

\begingroup
\renewcommand{\thefootnote}{*}
\footnotetext{Equal contribution.}
\endgroup

\lhead{}
\cfoot{}
\lfoot{\small Preprint.}

\begin{abstract}
Mechanistic interpretability (MI) aims to explain a model's behaviour through analyzing its internal computations; circuit-based explanations aim to isolate these computations with compact subnetworks validated by ablating the rest of the model. We show that circuits validated this way may fail to recover the underlying mechanism of the model's behaviour by closely reproducing its successful decisions while failing to account for most of its errors. Such explanations should account for the model's particular errors as well as its successes. We evaluate this requirement by measuring exact answer agreement separately on model successes and failures, across circuit sizes and ablation settings, on \textsc{IOI}, \textsc{Docstring}, and six model--task settings from the Mechanistic Interpretability Benchmark. We discover that many tested circuits closely replicate correct behaviour while missing most of the model's errors. On indirect object identification (\textsc{IOI}) for GPT-2 small, under mean ablation, the manual circuit and tested automated circuits, including one trained against the model's full output distribution, agree with the model on 97.3-99.5\% of prompts it answers correctly but only 11.4--41.7\% of errors. An \textsc{IOI} case study shows that lost errors are recoverable by restoring omitted attention-heads which raise error reproduction from 14.2\% to 75.1\% on a separate held-out set with 0.41 percentage point decrease on correct agreement, exceeding matched random extensions and scalar-biased control. Intervention traces show how omitted computations produce specific wrong answers for a reproducible subset of errors. In all, these findings show circuits can preserve task success without adequately explaining model's failures, and support exact error reproduction as a necessary, but not sufficient, test of circuit-based explanations of model behaviour. 
\end{abstract}


\section{Introduction}
\label{sec:introduction}

Mechanistic interpretability (MI) is a post-hoc interpretability method that aims to understand the underlying mechanisms of neural networks by reverse engineering them into human-understandable algorithms and circuits \citep{olah2020zoom}. By connecting computation units of the network and forming subgraphs, circuits attempt to explain the behaviour of the full model on specific tasks. These tasks formalize specific language-model behaviours as problems expressed as next-token prediction, such as identifying the recipient of an action or completing an arithmetic expression. Manual analysis \citep{wang2023interpretability, hanna2023how} and automated-circuit extraction methods \citep{conmy2023towards, syed-etal-2024-attribution} identify the components and connections responsible for these behaviours.

Prior work evaluates these circuits through faithfulness: how closely a circuit preserves the full model's behaviour under a metric and specific intervention \citep{wang2023interpretability, hanna2024have}. Common evaluations compare task scores, like average logit difference, or measure divergence between the model and circuit output distributions \citep{conmy2023towards}. These evaluations are popular for the identification of compact circuits and improvements in automated extraction. 

MI seeks to explain the computations underlying a model's behaviour, including the computations that produce mistakes. A circuit is a proposed account of that behaviour, therefore, it should be evaluated on inputs where the model is wrong as well as those where it is correct \citep{jacovi-goldberg-2020-towards}. We ask: \textit{to what extent do existing circuit-analysis methods account for naturally occurring model errors?}

Explaining the model's errors matters because an explanation of how the model succeeds on a task does not necessarily explain why it fails. A circuit that produces the correct answer when the model is wrong may capture useful task computation while ignoring influences that determine the model's real choice. Such failures limit its use as diagnostic tools: an explanation that removes the exact failure cannot, by itself, identify why the failure happened. Therefore, error reproduction tests a distinct scope of a circuit's explanatory ability, beyond its ability to perform the task correctly. 

We find that evaluated manual circuits and most tested automated extraction methods inadequately support model errors, often only preserving most of the correct behaviour. We characterize these failure modes and settings in which error reproduction succeeds, and use circuit completion and causal interventions on IOI to investigate the mechanisms involved. Our contribution covers the systematic evidence about the ability of existing circuit-analysis methods to account for the full mechanism of models.

\section{Preliminaries}
\label{sec:preliminaries}

\subsection{Tasks and Circuits}

\textbf{Tasks. } A \emph{task} is fixed by MI researchers as an observable behaviour under the model. It contains three parts: a distribution over prompts $x$; a \emph{labeling rule} $y(x)$ that fixes the correct answer for every prompt; and a measure of success. Because circuits are evaluated by patching in activations from another input, each prompt is replaced according to the specific ablation rule. 

\textbf{Circuits. } A circuit $\mathcal{C} \subseteq \mathcal{M}$ is a subgraph of the computation graph \citep{wang2023interpretability}. We evaluate it together with a patching setting specifying which activations or edge contributions outside the circuit are replaced and how their replacement values are obtained. Resample ablation uses activation from a counterfactual example, mean ablation uses averages over specified examples, and optimal ablation uses learned replacement constants. Retained computation is run forward on $x$. We define $\mathcal{C}(k \mid x)$ for the logit on candidate $k$ under the stated patching setting. 

\subsection{Faithfulness Metric}
Following \citet{heimersheim2023circuit}, who evaluate circuits by the proportion of prompts in which it predicts the correct argument, which \citet{miller2024transformer} classifies as a faithfulness metric, we similarly score predictions by first defining success as a computation over a \emph{candidate set}
$\mathcal{K}(x)$ of single-token answers with $y(x) \in \mathcal{K}(x)$.
Rather than free generation, every token is a ranked choice among candidates. These definitions apply to both the full model $\mathcal{M}$ and any circuit $\mathcal{C}$. The model's \emph{choice} on $x$ is
\begin{equation}
  \hat y_{\mathcal{M}}(x) \;=\; \argmax_{k \in \mathcal{K}(x)} \mathcal{M}(k \mid x),
  \label{eq:choice}
\end{equation}
The \emph{margin} of the label against its strongest candidate is
\begin{equation}
  m_{\mathcal{M}}(x) \;=\; \mathcal{M}\!\left(y(x) \mid x\right)
  \;-\; \max_{k \in \mathcal{K}(x) \setminus \{y(x)\}} \mathcal{M}(k \mid x).
  \label{eq:margin}
\end{equation}
\textbf{Error. } For disagreements between the label $y(x)$ and the model's choice  $\hat y_{\mathcal{M}}(x)$, we call this instance an \emph{error} of $\mathcal{M}.$ We define:
\begin{equation}
  \mathcal{E}(\mathcal{M})
  = \bigl\{x : \hat y_{\mathcal{M}}(x)\neq y(x)\bigr\},
  \qquad
  \varepsilon(\mathcal{M})
  = \Pr_{x\sim\mathcal{D}}\!\left[x\in\mathcal{E}(\mathcal{M})\right].
  \label{eq:error}
\end{equation}
for the error set and the error rate respectively. A negative answer margin implies an error and a positive margin implies a correct prediction. At zero, we use a deterministic tie rule (i.e. lowest token ID for IOI and docstring, refer to Appendix ~\ref{app:mib_details}, ~\ref{app:ioi_terminology} for MIB and IOI case study).

\textbf{Agreement. } We score the circuit against the model's choice $\hat y_{\mathcal{M}}(x)$ rather than the ground truth $y(x)$. This allows us to focus on whether $\mathcal{C}$ reproduces
$\mathcal{M}$, not whether $\mathcal{C}$ matches the task's correct $y(x)$. Let
$a(x) = \mathbb{1}\bigl[\,\hat y_{\mathcal{C}}(x) = \hat y_{\mathcal{M}}(x)\,\bigr]$,
and split it by the stratum $x$ falls in:
\begin{equation}
  A_{\mathrm{ok}} = \mathbb{E}\!\left[\,a(x) \mid x \notin \mathcal{E}(\mathcal{M})\,\right],
  \qquad
  A_{\mathrm{err}} = \mathbb{E}\!\left[\,a(x) \mid x \in \mathcal{E}(\mathcal{M})\,\right],
  \qquad
  \Delta = A_{\mathrm{ok}} - A_{\mathrm{err}}.
  \label{eq:gap}
\end{equation}
We call agreement on model-correct prompts correct agreement, \(A_{\mathrm{ok}}\), and agreement on model-error prompts error agreement, \(A_{\mathrm{err}}\).

\subsection{Benchmarks}
We use three major task families: \textsc{IOI} \citep{wang2023interpretability}, \textsc{Docstring} \citep{heimersheim2023circuit}, and \textsc{MIB} \citep{mueller2025mib}. 

\paragraph{IOI. }
We generate prompts following the standard \textsc{IOI} protocol, using 15 template families in both ABBA and BABA order. We sample distinct names and fill place and object slots from the library vocabularies, restricting to single-token entries. Each prompt contains two names and repeats one as the subject (S) where the correct continuation is the other name, the indirect object (IO). We remove the final answer from the generated sentence and evaluate GPT-2 Small over $\mathcal{K}(x)=\{\mathrm{IO},\mathrm{S}\}$, with
$y(x)=\mathrm{IO}$. For counterfactual ablations, we use same-template ABC reference prompts containing three distinct names. We evaluate both resample ablation and mean ablation. 

\paragraph{Docstring.}
From the public ACDC docstring benchmark, we generate synthetic Python function signatures and partially completed docstrings. The argument names and description words are randomly sampled from fixed vocabularies. The task is defined as predicting the next argument name in the docstring in the order established in the function signature. We evaluate a four-layer attention-only transformer over the benchmark's 14 single-token candidates: the correct argument and 13 distractors. We generate counterfactual prompts from the generator's \texttt{random\_random} corruption, which replaces argument names in both the signature and docstring to break their correspondence. 

\paragraph{MIB.} 
We use the published prompts and dataset splits from \textsc{MIB}, specifically ARC-Easy, ARC-Challenge, and arithmetic subtraction in our main evaluations. ARC consists of multiple-choice science questions where the answer letter option labels as candidates \citep{clark2018thinksolvedquestionanswering}. For subtraction, candidates are single-token integers from 0 to 99. We use the supplied symbol counterfactuals for ARC and random-operand counterfactuals for subtraction. We omit Addition and synthetic MCQA from the experiments because they contain only seven and five full-model errors on the public test splits, respectively, yielding loose error-agreement estimates.

We do not include \textsc{Greater-Than} year-span benchmark as \citet{hanna2023how} report 100\% correct top-1 performance; similarly, for well-posedness, we restrict each task to models whose answers is contained in a single token i.e subtraction is ran on Llama-3.1 8B only (as Gemma-2 and Qwen-2.5 tokenize numbers by digit so answers fall out of candidate sets \ref{eq:choice}). Full details are reported in Appendix ~\ref{app:mib_details}.

\section{Agreement Analysis}
\label{sec:circuit_model_agreement}

\begin{figure}[t]
\centering
\includegraphics[width=1.0\textwidth]{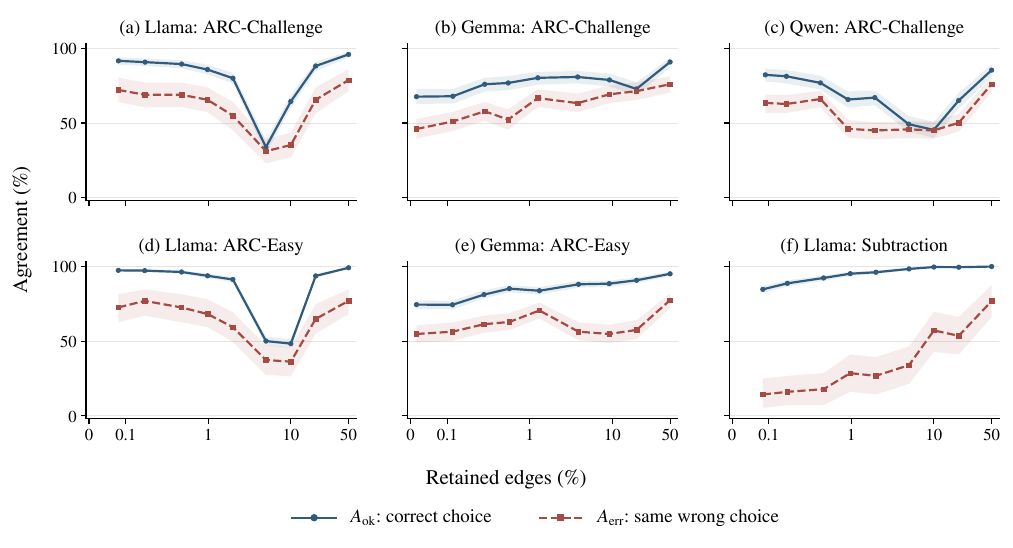}
\caption{Preservation of correct and incorrect model prediction across circuit sizes. Circuits are extracted on the full public-training split and evaluated on the public-test split. Shading indicate the point-wise 95\% bootstrap confidence intervals. Specific models are: Llama-3.1 (8B), Gemma-2 (2b), Qwen-2.5 (0.5B).}
\label{fig:agreement-gap}
\end{figure}

Table~\ref{tab:main} compares $A_{\mathrm{ok}}$ and $A_{\mathrm{err}}$ (Eq.~\ref{eq:gap}) under mean and paired resample ablation for \textsc{IOI} and \textsc{Docstring}. Figure~\ref{fig:agreement-gap} reports the \textsc{MIB} evaluations under counterfactual activation replacement values. We also evaluate optimal ablation \citep{li2024optimal}, whose replacement values are learned to minimize KL divergence from the full model's output distribution. For both \textsc{IOI} and \textsc{Docstring}, ACDC \citep{conmy2023towards} and our EAP-IG configuration \citep{hanna2024have} use KL divergence from the full model's output distribution as their discovery objective. Edge Pruning, evaluated on \textsc{IOI}, also uses this model-matching KL loss alongside its sparsity objective. These objectives target the model's output distribution, including on examples where it is wrong, without explicitly rewarding the ground-truth answer.

\begin{table*}[t]
\centering
\caption{Normalized task faithfulness and exact agreement with
$\mathcal{M}$ under argmax over the benchmark candidate set
$\mathcal{K}(x)$. EAP-IG and Edge Pruning use validation-selected
recovery circuits; manual and ACDC circuits are reference checkpoints.}
\label{tab:main}
\begingroup
\small
\setlength{\tabcolsep}{3pt}
\renewcommand{\arraystretch}{1.08}
\begin{adjustbox}{max width=\textwidth}
\begin{tabular}{@{}lllrrrr@{}}
\toprule
Circuit / model & Size & Ablation & $F$ &
\multicolumn{2}{c}{Agreement (\%)} & $100\Delta$ (pp) \\
\cmidrule(lr){5-6}
& & & & $A_{\mathrm{ok}}$ & $A_{\mathrm{err}}$ & [95\% CI] \\
\midrule

\multicolumn{7}{@{}l}{\textit{IOI}} \\
\addlinespace[2pt]

Manual reference & 26 heads & Mean
& 0.956 & 99.5 & 15.2 & $84.3\;[82.0,86.6]$ \\
& & Resample (ABC)
& 0.644 & 91.2 & 40.6 & $50.5\;[47.1,54.0]$ \\
& & OA (KL)
& 0.906 & 99.9 & 10.0 & $90.0\;[87.9,91.9]$ \\
\addlinespace[2pt]

ACDC ($\tau=0.003$) & 402 edges & Mean
& 1.167 & 97.3 & 11.8 & $85.5\;[83.3,87.6]$ \\
& & Resample (ABC)
& 0.887 & 91.5 & 24.8 & $66.8\;[63.8,69.7]$ \\
& & OA (KL)
& 0.869 & 99.7 & 8.1 & $91.6\;[89.8,93.3]$ \\
\addlinespace[2pt]

EAP-IG-KL & 6,498 edges & Mean
& 0.987 & 99.1 & 41.7 & $57.4\;[54.2,60.6]$ \\
& & Resample (ABC)
& 0.905 & 97.8 & 51.4 & $46.4\;[43.1,49.7]$ \\
& & OA (KL)
& 0.985 & 99.8 & 44.0 & $55.8\;[52.5,59.1]$ \\
\addlinespace[2pt]

Edge Pruning & 256/268/268 edges & Mean
& 1.486 & 98.3 & 11.4 & $86.9\;[85.2,88.5]$ \\
& & Resample (ABC)
& 1.002 & 93.7 & 17.6 & $76.0\;[73.7,78.3]$ \\
& & OA (KL)
& 0.971 & 99.2 & 15.2 & $84.0\;[82.1,85.8]$ \\

\midrule
\multicolumn{7}{@{}l}{\textit{Docstring}} \\
\addlinespace[2pt]

Manual reference & 8 heads & Mean
& 1.017 & 77.1 & 35.8 & $41.3\;[39.4,43.1]$ \\
& & Resample
& 0.765 & 47.2 & 38.9 & $8.2\;[6.2,10.2]$ \\
& & OA (KL)
& 0.998 & 81.6 & 42.5 & $39.1\;[37.3,41.0]$ \\
\addlinespace[2pt]

ACDC ($\tau=0.005$) & 129 edges & Mean
& 0.729 & 67.8 & 53.3 & $14.4\;[12.4,16.4]$ \\
& & Resample
& 0.918 & 71.7 & 56.3 & $15.4\;[13.4,17.3]$ \\
& & OA (KL)
& 1.006 & 83.3 & 41.5 & $41.9\;[40.0,43.7]$ \\
\addlinespace[2pt]

ACDC ($\tau=0.02$) & 52 edges & Mean
& 0.990 & 74.8 & 37.2 & $37.6\;[35.7,39.5]$ \\
& & Resample
& 0.866 & 59.0 & 42.0 & $17.1\;[15.1,19.1]$ \\
& & OA (KL)
& 0.898 & 78.9 & 39.1 & $39.8\;[37.9,41.7]$ \\
\addlinespace[2pt]

EAP-IG-KL & 500 edges & Mean
& 0.651 & 65.6 & 69.4 & $-3.8\;[-5.7,-1.9]$ \\
& & Resample
& 0.842 & 66.3 & 66.3 & $0.0\;[-2.0,2.0]$ \\
& & OA (KL)
& 1.043 & 89.9 & 60.0 & $29.9\;[28.2,31.7]$ \\

\bottomrule
\end{tabular}
\end{adjustbox}

\par\smallskip
\begin{minipage}{\textwidth}
\footnotesize
EAP-IG and Edge Pruning use the smallest evaluated non-full
circuit with normalized validation-resample faithfulness
$F_{\mathrm{val}}\geq0.85$, where
$F=(\bar m_{\mathcal C}-\bar m_{\emptyset})/
(\bar m_{\mathcal M}-\bar m_{\emptyset})$ with graph- and
ablation-specific empty controls. Edge Pruning lists three fitted-seed sizes and averages their agreement.
Intervals use 10,000 paired stratified bootstrap resamples,
retain sampling weights, and condition on fitted masks.

\end{minipage}
\endgroup
\end{table*}

\paragraph{High task faithfulness can conceal poor error reproduction.}
Prior work has shown that average circuit evaluations can miss substantial disagreements on individual inputs \citep{bos2024adversarial}, and has distinguished task performance from reproducing model behaviour \citep{miller2024transformer}. Similarly, our results support a systematic deficit on the model's own errors. On IOI under mean ablation, the manual circuit, ACDC, EAP-IG-KL, and Edge Pruning \citep{bhaskar2024edge} preserve 97.3\%--99.5\% of the correct choices but reproduce only 11.4\%--41.7\% of errors. The manual circuit has normalized mean logit-difference faithfulness of $F = 0.956$, yet reproduces only $15.2\%$ of the model's errors. This gap also happens in circuits discovered with model-matching KL objectives. However, error reproduction still varies with the evaluation setting; EAP-IG reaches $51.4\%$ under resampling at 6,498 edges.

\paragraph{Smaller gaps can reflect either better error reproduction or lost correct behaviour.}
For the \textsc{Docstring} manual circuit, resampling reduces the gap from $41.3\%$ to $8.2\%$, mainly due to the correct agreement falling from $77.1\%$ to $47.2\%$. In contrast, \textsc{Docstring} EAP-IG instead reproduces $69.4\%$ of errors under mean ablation which exceeds its $65.6\%$ correct agreement, suggesting that an extracted circuit can preserve substantial performance on both behaviours.

\paragraph{Discovery error exposure and weighting provide only partial recovery.}
In the six \textsc{MIB} settings, 187--849 model errors appear in the public-training splits, representing 4.9--41.1\% of the discovery examples. On Gemma ARC-Challenge, 430 of the 1,119 discovery examples ($38.4\%$) are errors, yet across most edge budget sizes, there's a sustained gap. For instance, at 1\% edge budget, the circuit preserves 80.3\% of correct choices and only reproduces 54.4\% of the model's wrong answers. Thus, the disparity cannot be soley attributed simply to errors being rare during discovery as it still persists with roughly 60:40 correct to error split. Nevertheless, we test whether increasing the weight assigned to errors during discovery improves their reproduction in a separate ACDC follow up on IOI: while assigning existing errors 50\% of the total discovery KL weight raises average exact-error agreement under mean ablation from 9.3.\% to 37.0\%, it causes correct agreement to fall from 98.7\% to 91.0\%. Reweighting therefore improves error reproduction, but still leaves a significant agreement gap while incurring a non-trivial correct-case cost. 

\paragraph{Matching margins diminishes some disparities.}
Model errors may be concentrated at small margins \citep{hendrycks2017a}, increasing the probability that smaller margins make their predictions more sensitive to differences between the circuit and model logits. Therefore, we match correct and error test examples on the absolute model margin between the highest and second highest candidate logits computed from the full model over the candidate set $\mathcal{K}(x)$, within each setting for \textsc{IOI}, \textsc{Docstring}, and \textsc{MIB}. Margin matching does decrease several MIB ARC agreement gaps. For Llama ARC-Challenge at 2\% edge budget, correct agreement is $80.0\%$ across all correct examples, but $46.7\%$ among the matched correct examples, while exact-error agreement remains at $54.9\%$. The narrowed gap shows lower preservation of correct examples within comparable margins. Some disparities do exist like Gemma ARC-Easy retaining gaps of 15.3--21.2 \% at the 5\%, 10\%, and 20\% budgets. Therefore, several agreement gaps are smaller on margin-matched examples, with the remaining gaps depending on the setting and circuit size. Positive gaps also persist after margin matching in all \textsc{IOI} and most \textsc{Docstring} configurations reported in Table~\ref{tab:main}. Full results for IOI, Docstring, and MIB are provided in the Appendix~\ref{app:margin_matching}.


\paragraph{Additional recovery attempts have mixed effects.}

We also test discovery sets with increased distinct model errors, error-conditioned activation replacements, and post-hoc calibration of circuit logits and decision thresholds. In the ACDC IOI follow-up, using a discovery split of 500 error and 500 correct examples results in a $A_{err}$ of 44.2\% and $A_{ok}$ of 91.4\%. Error-conditioned mean and resample replacements also improve IOI $A_{err}$, but simply shifting the output logits toward the subject name reproduces at least as many errors. Matching the model's overall error rate does not consistently recover its particular wrong answers, while more aggressive threshold calibration can narrow the gap by sacrificing correct performance.

\section{Case Study: why the manual reference circuit misses model errors}
\label{sec:ioi_case_study}

We next show in an IOI case study that restoring selected omitted components can substantially improve exact-error agreement while largely preserving correct behaviour, and trace the computations underlying these recoveries.

\subsection{Understanding missing errors}
\label{sec:questions}
We organize our case-study around three primary questions:  

\paragraph{Q1: Which omitted components contribute to missing errors, and how?}
We test whether restoring selected components reproduces the model's exact wrong answers, and whether they act through separate mechanisms or modify existing circuit computations. 

\paragraph{Q2: Can error recovery improve while largely preserving correct agreement?}
We analyze whether restricted extensions improve error agreement with limited losses on correct examples, reporting both changes relative to the reference circuit and the number of added components. 

\paragraph{Q3: How concentrated and stable are the missing contributions?}
We examine whether errors share omitted contributions and whether their effects persist across ablation settings and held-out tests.

\begin{figure}[t]
\centering
\includegraphics[width=1.0\textwidth]{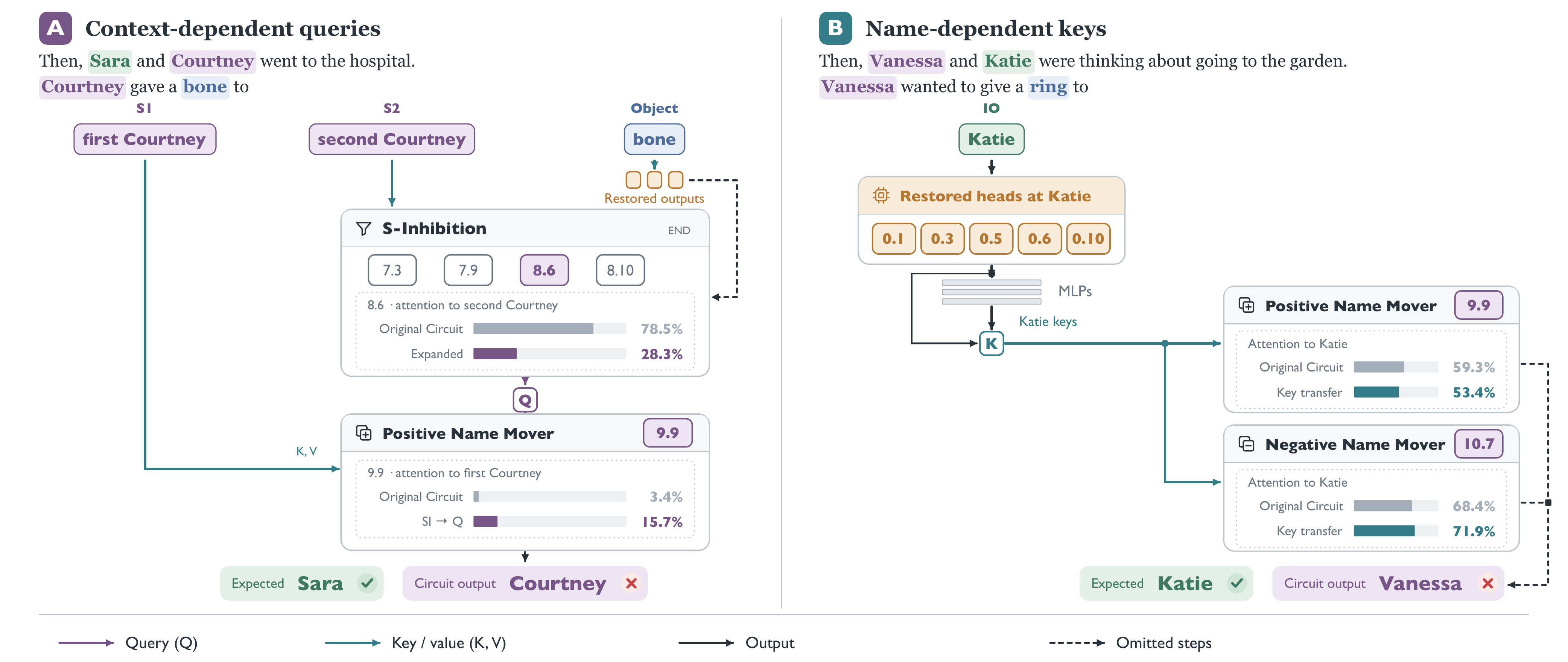}
\caption{\textbf{Two omitted computations recover model errors:} A) Restored contextual computation changes S-Inhibition and Name Mover queries, producing Courtney instead of Sara. B) Restored-name source computations changes keys, reducing the retrieval of Katie by a Positive Name Mover and increasing it by a Negative Name Mover, both favouring Vanessa over Katie.}
\label{fig:omitted_computations_graph}
\end{figure}

\paragraph{Experimental setup. }

We use GPT-2 small and \citet{wang2023interpretability}'s published IOI reference circuit $C_0$ and, similarly, search for attention heads $h$ at specific positions. We search over following positional token rules: \textsc{END} (the last input position), \textsc{IO} (the recipient's name), \textsc{S1} and \textsc{S2} (the subject's first and second occurrences), \textsc{S1 + 1} (the position after the subject's first occurrence), \textsc{BOS} (the beginning of sequence position). The error-recovery experiment in Section~\ref{sec:4.2} uses an expanded search space, described below. To further match the IOI study\citep{wang2023interpretability}, we scope our investigation on the attention heads while ablating with mean knockouts, and we do not intervene on any other components. We provide definitions of IOI-specific terminology and additional details of the reference circuit in Appendix~\ref{app:ioi_terminology}.

The reference circuit contains several functional head groups \citep{wang2023interpretability}. Positive Name Movers attend to earlier name tokens and increase those names' output logits; Negative Name Movers decreases them. Backup Name Movers perform a similar name-copying function when primary Name Movers are ablated. S-Inhibition heads use information about the repeated subject to influence the Name Mover queries, helping them avoid copying the subject.

We search for a compact extension of $C_0$ that more faithfully reproduces the full model's ($\mathcal{M}$) predictions on examples where $\mathcal{M}$ is incorrect, while preserving the agreement on examples it answers correctly.

Starting from $C_0$, we greedily expand the circuit using a gradient-based screening to identify head-position candidates, followed by exact binary-intervention evaluation. We guide our search with the following objective:
\begin{equation}
J(\mathcal{C})
=
A_{err}(\mathcal{C})
-
\alpha
\left[
A_{ok}(\mathcal{C}_0)-A_{ok}(\mathcal{C})-\epsilon
\right]_+
-
\lambda_P P_{\mathrm{added}}(\mathcal{C}),
\end{equation}
where $[z]_+ = \max(z, 0), \epsilon$ is the tolerated decline in correct agreement, and $P_{\mathrm{added}}(\mathcal{C})$ counts added head-position nodes beyond $\mathcal{C}_0$,

Our objective reflects two standard principles in circuit extraction: faithfulness and minimality \citep{wang2023interpretability, conmy2023towards}. The recovery term pushes towards recovering $\mathcal{M}$ on $\mathcal{E}(\mathcal{M})$, while the agreement penalty penalizes the recovery from degrading the initial correct performance for $C_0$. The sparsity terms favour compact extensions, with separate costs for introduce new heads. We provide the full algorithm and hyperparameters in the Appendix \ref{app:ioi_terminology}.

\subsection{Error Recovery On Fresh Data}
\label{sec:4.2}

For this experiment, we expand the search space to 25 positional rules across all 144 attention heads, yielding 3,574 supported head–position additions beyond the reference circuit \(C_0\). The selected extension, $C_*$, adds 63 position--head nodes, with approximately 61.54 additional outputs per prompt on average. 
The final selected extension adds 63 rules. On a frozen 26,000-example test set containing 225 $\mathcal{M}$ errors, the reference circuit reproduces only $14.22\%$ errors, whereas the selected extension reproduces $75.11\%$. This is with only a correct agreement drop of $0.41$ percentage points from $99.34\%$ for the reference to $98.93\%$ for the extension. Despite the strong class imbalance towards correct examples, the extension still gains more agreements in absolute terms than it loses, recovering 137 additional model errors while losing 106 correct examples. $C_*$ also exceeds all structurally matched random extensions (error agreement $17.33$--$41.33\%$, median $29.33\%$). A scalar subject-bias applied to $C_0$, fitted on validation correct cases to match $C_*$'s correct agreement, reached only $20.44\%$ exact error agreement on this test set at nearly identical correct agreement ($98.93\%$).

\textbf{This answers Q2 affirmatively on this evaluation:} a restrictived position-specific extension is able to recover a significant gain of model failures that are not reproduced by the published reference circuit, while largely preserving its behaviour on examples the model answers correctly. 

\subsection{How omitted computations change which name wins}

\textbf{To address Q1, } given the existing subject-inhibition (S-inhibition heads) and name-writing (Name Mover heads) components, our analysis identifies two ways in which omitted inputs alter its behaviour: \textbf{contextual information changes the queries used to select names, while early name-dependent writes change the keys through which the correct name is retrieved.} Retained heads and MLPs can amplify or oppose both of these effects. 

\paragraph{Mechanistic analysis.} We study all 137 new reproduced errors, denoted as $G.$ Like \citep{wang2023interpretability}, we use activation patching and path patching to test whether selected activation changes from the expansion make $C_0$ reproduce the model's wrong answer. Conversely, we also patch reference-circuit activations into $C_*$ and $\mathcal{M}$ to test whether these replacements restore the correct answer. To find the context that is responsible, we patch activations from counterfactual prompts with altered object words, name identities, or name positions, while keeping the receiving prompt unchanged. We compare the logits of the indirect object (IO) and subject (S). \textit{Recovery} refers to reproducing the model's preference for the incorrect subject name; \textit{repair} means restoring the preference for the correct IO name. 

\paragraph{Contextual inputs effect subject inhibition. } Existing S-inhibition heads read information about the repeated subject at \textsc{S2} and influence which names the Name Mover heads select. Restored object-related inputs change how the inhibition heads read this information. In the traced computation, the resulting query changes increase downstream attention to the subject, causing the retained Name Movers to copy the wrong name, illustrated by Figure ~\ref{fig:omitted_computations_graph}. To test this mechanism, we isolate the difference between $C_*$ and $C_0$ at the part of the inhibition output obtained by attending to \textsc{S2}. Patch patching this changed inhibition contribution into the Name Mover queries of $C_0$, with reference keys and values fixed causes it to select the wrong subject on 120 of the 137 error prompts. Thus, retained heads can produce the wrong answer when supplied with altered inhibition signals. 

\paragraph{Testing the influence of the object content.} We construct corresponding prompts in which the object word is replaced with "snack." For instance, "gave a bone to" becomes "gave a snack to", while keeping the remaining text untouched. These altered prompts supply the replacement content for the information retrieved from the object position by the added attention heads in $C_*$. The circuit still operates on the original prompt, and the attention weights at the substituted outputs remain unchanged. This isolates the influence of object-derived information inside the circuit. The substitution of this object restores the correct IO answer on 107 of the 137 prompts. Then, we repeat it while holding the S-Inhibition outputs at the same values from $C_*$ before the substitution. Under this constraint, 68 of those corrections disappear. This suggests an object-dependent influence on subject inhibition, however, this does not disclose which exact semantic feature of the object causes it. 

\paragraph{Early name processing affects retrieval.}

A second omitted computation changes the representation of the correct name before later heads retrieve it. The extension restores the outputs of five first-layer heads: $0.1$, $0.3$, $0.5$, $0.6$, and $0.10$, at the IO position; $\ell.h$ denotes the attention head $h$ in layer $\ell$, with both indices starting at zero. Their outputs combine with downstream MLP transformations to change the IO representation used to compute the Name Movers' keys, shown in ~\ref{fig:omitted_computations_graph}B. After activation patching the Name Movers' keys at the IO position with their values from $C_*$ on the same prompt, the intervention reproduces the model's wrong subject answer on 35/137 examples. Restoring the keys only at either subject occurrence, or only at the remaining positions, reproduces none. In the measured route, the changed keys reduce IO retrieval by Positive Name Movers and increase it by Negative Name Movers, whose IO-source values oppose the correct name in the IO minus subject margin. 

\paragraph{Concentration and interactions (Q3)}
Path patching the S-Inhbition heads' \textsc{S2} contribution into $C_0$'s Positive, Negative, and Backup Name Mover queries reproduces 120/137 newly recovered errors, showing a recurring contribution across examples. Additionally, the extension's performance persists under resample ablation: error agreement increases from $40.89\%$ to $60.89\%$, and the correct agreement from $91.41\%$ to $97.67\%$. Therefore, 
recovery involves re-recurring interactions with retained computation and is observed under both replacement settings.

\section{Recommendations}
\label{sec:recommendations}
From our findings in Section ~\ref{sec:circuit_model_agreement} and ~\ref{sec:ioi_case_study}, we recommend the following checks to test whether a circuit reproduces the model’s choices on both model-correct and model-incorrect prompts, based on the stated circuit execution and ablation rule. However, this alone does not establish whether the circuit recovers the model’s underlying mechanism \citep{wang2023interpretability}.  

\textbf{Build Stratum.} For each data split, partition the prompts based on whether the full model correctly answers the task. The strata is then fixed across all evaluated circuits, sizes and ablation methods.

\subsection{Raw Metrics for Extracted Circuit}
Report $A_{ok}, A_{err}$ together with their difference $\Delta$ as defined in Eqn~\ref{eq:gap}. If there are no model errors, $A_{err},\Delta$ is undefined. Note that specifically for $x \in \mathcal{E}(\mathcal{M})$ where $\hat y_{\mathcal{M}}(x) \neq y(x)$: $a(x)=1$ requires the circuit to return \emph{the same incorrect} answer the model returns. A circuit returning the correct answer for the task when the full model was wrong or a different incorrect answer from the full model counts as a disagreement. As an additional diagnostic for future circuit evaluations, to distinguish these two types of disagreement, we define $B_{\mathrm{err}}(\mathcal{C})$, the fraction of model-error prompts on which the circuit is also incorrect:
\begin{equation}
    B_{\mathrm{err}}(\mathcal{C}) \;=\; \mathbb{E}\!\left[\, \mathbb{1}\!\left[\hat y_{C}(x) \neq y(x)\right] \;\middle|\; x \in \mathcal{E}(\mathcal{M}) \,\right]
\end{equation} 
The shares on which the circuit selects the correct answer and a different incorrect answer are \(1-B_{\mathrm{err}}(\mathcal{C})\) and \(B_{\mathrm{err}}(\mathcal{C})-A_{\mathrm{err}}\), respectively. For binary tasks, \(B_{\mathrm{err}}(\mathcal{C})=A_{\mathrm{err}}\).

\subsection{Evaluation Across Circuit Sizes}
Evaluation at a single circuit size may not be sufficient in accounting for how these numbers change as circuit size changes. Following \citet{mueller2025mib}'s use of a set containing proportions of components, we recommend evaluating agreement across a specified sequence of circuit sizes. Crucially, we track agreement on each stratum separately across sizes, because a high average logit-difference score does not necessarily imply circuit matches the model's individual predictions \citep{miller2024transformer}.  

\textbf{Size sequence.} When a discovery method provides a ranking or budget sequence, specify the evaluated sizes before final evaluation and report the retained component counts and their units. For our MIB experiments, we evaluate circuits at retained-edge fractions, $s\in\{.001,.002,.005,.01,.02,.05,.1,.2,.5,1\}$ and report empty circuit ($s=0$) control. A circuit with no defined sequence can be reported as individual points.

\textbf{Stratified curves.}
Plot $A_{\mathrm{ok}}(s)$ and $A_{\mathrm{err}}(s)$ evaluated on $\mathcal{C}_s$ against $s$. 
Their difference, $\Delta(s)$, shows whether $A_{err}$ is lower at every non-full model size or only at the size we happened to pick. A smaller gap can reflect improved error agreement, but could also mean reduced correct agreement. 

\paragraph{Reporting.}
In short, our main recommendation is to separately report agreement on model-correct and model-error prompts, requiring the same incorrect answer on the error stratum in particular. To make these results interpretable, report the evaluated split, the number of prompts in each stratum, circuit size and ablation rule, since circuit evaluations depend on these methodological choices ~\citep{miller2024transformer}.

\section{Related Work}
Manual circuit analyses aim to discover compact explanations of model behaviour \citep{wang2023interpretability}, while automated methods like ACDC and EAP-IG automate the identification of relevant components \citep{conmy2023towards, hanna2024have}. Evaluating these explanations still remains as a challenge. \citet{miller2024transformer} show that circuit faithfulness depends greatly on ablation methodology, and \citet{shi2024hypothesis} develop statistical tests of behavioural preservation, localization, and minimality. \citet{mueller2025mib} separates components that improve task performance from those with any measurable influence, including negative effects, and evaluates circuit discovery across sizes. \citet{yaniv2025arithmetic} analyze arithmetic heuristics and link their failures with reduced support for correct-answer logits, while \citet{bertolazzi-etal-2025-validation} show that models often accept incorrect arithmetic solutions because they check whether the stated numbers agree with each other rather than if the calculation is correct. These works provide mechanistic accounts of specific model failures. Our work systematically tests whether existing circuit explanations reproduce their models' wrong answers, revealing frequent shortfalls relative to the correct-case preservation and tracing omitted computations that recover a subset of those errors. 


\section{Limitations and Conclusion}

\paragraph{Limitations.}
Our findings spans \textsc{IOI}, \textsc{Docstring}, and \textsc{MIB} across multiple models, extraction methods, circuit sizes, and ablation settings, but our conclusions are scoped to these tested settings and their benchmarks. Our detailed case study concerns one IOI reference circuit and its found extension; 
it illustrates specific recovered errors without establishing that the same mechanisms generalize to other circuits or tasks. Although several recovery strategies were tested and demonstrations indicate that significant recovery is possible, we neither exhaust all possible remedies nor establish a generalized solution for recovering omitted computations across all circuit settings while consistently preserving correct behaviour. 

\paragraph{Conclusion.}
Explaining a model's behaviour requires accounting for its failures as well as its successes. The tested manual circuits and most automated extraction methods we evaluate often fail this requirement, reproducing correct predictions substantially better than the model's particular wrong answers. Our IOI case study demonstrates that missing errors can be recovered and traces how specific missed computations produce these specific wrong answers. Therefore, we recommend evaluating exact wrong-answer reproduction alongside correct-answer preservation as a necessary test of circuits intending to explain model behaviour.

\clearpage

\bibliography{iclr2027_conference}
\bibliographystyle{iclr2027_conference}

\clearpage
\appendix

\section{Experiment Details: IOI and Docstring}

\subsection{Models, prompts, and evaluation populations}
\label{app:a1}

All \textsc{IOI} experiments use the frozen \textsc{GPT-2} small checkpoint (12 layers, 12 attention heads per layer). \textsc{Docstring} uses \textsc{NeelNanda/Attn\_Only\_4L512W\_C4\_Code}, a four-layer, eight head per layer, attention only model. In these computations, we use float32 with TF32 disabled; model weights are fixed throughout extraction and replacement fitting.

\paragraph{IOI Construction.}The generator uses 15 template families with ABBA and BABA name order, resulting in 30 template/order cells. We filter names, places, and objects to single tokens. Two distinct names fill each clean prompt, with the repeated name representing the subject (S) and the other name as the indirect object (IO). For ablation, we sample a corresponding prompt from the same template using three distinct names, producing an ABC prompt in which the repetition structure is removed. The lengths of $x'$ and $x$ token lengths must match. Therefore, $x'$ may change object and place words as well as names, instead of a names-only counterfactual. 

\paragraph{Docstring construction.} We use the public ACDC generator in its reST-style format. Each prompt contains a function definition with two prefix arguments, followed by three matching arguments and a suffix argument.  The docstring documents the first two matching arguments, and the task is to predict the third as the next entry. We use no docstring-prefix arguments, a three-word method description, two-word argument descriptions, and function arguments without default values. The next argument is scored against a fixed set of 14 benchmark candidate argument tokens (the correct argument and 13 alternatives). For ablations, $x'$ is from \texttt{random\_random} corrupted dataset used by ACDC, which independently randomizes the variable names in both the function definition and the docstring.

\paragraph{Coverage and seperation.}Our discovery pools contain 13,000 \textsc{IOI} examples and 3,000 \textsc{Docstring} examples which includes both correct and incorrect model predictions, with equal weight per example. Discovery, validation, and test sets are disjoint, including the prompts used for ablation.

For \textsc{IOI}, we evaluate the circuits on 10,884 prompts sampled from the 100,000 prompt test pool while retaining all model wrong answers and weighting results to account for the sampling. For \textsc{Docstring}, we evaluate circuits on all 10,000 test prompts.

\paragraph{Prediction and uncertainty.} Candidates are sorted by token ID before argmax, so on ties, the lowest token ID is selected. Correct/error classification is defined by this prediction. Exact error agreement counts the identical wrong candidate on every model error. 
The reported gap intervals use 10,000 percentile bootstrap draws, resampling within the three \textsc{IOI} sampling cells. For Edge Pruning, we average the three seed agreement indicators for each prompt before bootstrapping shared prompts.

\begin{table}[t]
\centering\small
\caption{IOI/Docstring data coverage. Discovery pools are used in full for current EAP-IG,
Edge Pruning where supported, and OA fitting.}
\label{tab:current-split-coverage}
\begin{tabular}{llrrr}
\toprule
Task & Split & Rows evaluated & Candidate errors & Candidate correct \\
\midrule
IOI & Discovery pool & 13,000 & 135 & 12,865 \\
IOI & Validation & 10,000 & 95 & 9,905 \\
IOI & Test census & 100,000 & 884 & 99,116 \\
IOI & Circuit-scored test sample & 10,884 & 884 & 10,000 \\
Docstring & Discovery pool & 3,000 & 1,027 & 1,973 \\
Docstring & Validation & 2,000 & 676 & 1,324 \\
Docstring & Test & 10,000 & 3,536 & 6,464 \\
\bottomrule
\end{tabular}
\end{table}

\subsection{Discovery Implementations}

\paragraph{Manual references.}
The \textsc{IOI} reference contains 26 attention heads at specific positions: previous-token heads at S1+1, induction and duplicate-token heads at S2, and the remaining heads at END. All other attention-head outputs are ablated with MLPs remaining active. We provide Appendix ~\ref{app:ioi_terminology} which lists the retained heads details and positions. For \textsc{Docstring}, the retained reference heads are: $0.2, 0.4, 0.5, 1.2, 1.4, 2.0, 3.0, 3.6.$ 

\paragraph{ACDC.} ACDC checks a candidate edge rmoval against the current circuits and commits to the removal when the increase in mean $D_{\mathrm{KL}}(p_{\mathcal M}\Vert p_{\mathcal C})$ is below its selected threshold \citep{conmy2023towards}. For both tasks, the loss compares full-vocabulary distributions at the answer position without using ground-truth labels. For \textsc{IOI}, we use an implementation built on AutoCircuit \footnote{\url{https://github.com/UFO-101/auto-circuit}}. For \textsc{Docstring}, we use the authors' ACDC implementation\footnote{\url{https://github.com/ArthurConmy/Automatic-Circuit-Discovery}} with redundant-edge removal and absolute-value thresholding disabled. These fixed circuits are evaluated on the held-out test sets.

\paragraph{EAP-IG.} EAP uses activation difference and local gradients to approximate edge interventions \citep{syed-etal-2024-attribution}. We use the input-integrated-gradient variant and circuit search of \citet{hanna2024have}.\footnote{
\url{https://github.com/hannamw/eap-ig}}. Attribution scores are computed over the complete discovery split using five interpolation steps from the corrupt to clean input embeddings. The objective is the full-vocabulary $D_{\mathrm{KL}}(p_{\mathcal M}\Vert p_{\mathcal C})$
at the answer position. Circuits are constructed backward from the output using absolute attribution scores, allowing for both positive and negative attributions to be picked up, followed by the removal of disconnected components. 

\paragraph{Edge Pruning.}
We use the authors' \textsc{IOI} implementation of Edge Pruning \citep{bhaskar2024edge}.\footnote{
\url{https://github.com/princeton-nlp/Edge-Pruning}}. This method learns a sparse circuit by using learnable gates on edges between the transformer components and optimizing these gates to preserve the model's behaviour while encouraging sparsity. We fit a range of sparsity targets using three random seeds and over the complete 13,000 example discovery set. Each fit runs for 3,000 updates with batch size of 32, learning rate of 0.8, and 2,500 step sparsity warmup. The separate node-loss term is disabled. Dropout is disabled and gradients are clipped at norm 1. The three
random seeds are 0, 1, and 2. The sparsity-target grid is
\[
\begin{split}
\{&0.2,0.4,0.6,0.8,0.9,0.94,0.945,0.95,0.955,0.96,0.965,0.97,\\
  &0.975,0.98,0.985,0.99,0.995,1,1.01,1.02,1.05,1.1\}.
\end{split}
\]
Targets above one are optimization targets, not realized sparsity
fractions. We discretize the final training checkpoint using the
implementation's thresholding procedure and report actual retained
edge counts. 

We don't evaluate Edge Pruning on \textsc{Docstring} as the official implementation does not support its attention-only architecture.

\subsection{Replacement activations and normalized faithfulness}

\begin{table}[t]
\centering\small
\caption{OA training for current recovery/reference rows. Selection uses minimum validation KL}
\label{tab:oa-fit-status}
\begin{tabular}{llrrrr}
\toprule
Task & Circuit & Epochs & Updates & Selected epoch \\
\midrule
IOI & Manual / ACDC / EAP-IG & 13 & 10,725 & 12 \\
IOI & Edge Pruning seed 0 & 12 & 9,900 & 9 \\
IOI & Edge Pruning seeds 1, 2 & 13 & 10,725 & 12\\
Docstring & Manual & 26 & 4,888 & 23 \\
Docstring & ACDC $\tau=.005$ & 26 & 4,888 & 23 \\
Docstring & ACDC $\tau=.02$ & 29 & 5,452 & 26 \\
Docstring & EAP-IG & 30 & 5,640 & 27 \\
\bottomrule
\end{tabular}
\end{table}

Mean IOI replacements are per-template, per-position averages of the source activations on 128 independent ABC donors. For \textsc{Docstring}, we generate a separate set of 1,024 prompt pairs using the same ACDC docstring generator and corruption procedure as the task dataset. We exclude duplicate prompts across this set and the discovery and evaluation sets. Mean replacements are position-specific averages of source activations over the corrupted prompts in this reference set. 

\textbf{Optimal ablation.} Following optimal ablation \citep{li2024optimal}, we replace each component outside the fixed circuit with a learned, input-independent constant. For each circuit, all constants are jointly learned to minimize the full-vocab $D_{\mathrm{KL}}(p_{\mathcal M}\Vert p_{\mathcal C})$ between the full model and the circuit. Each constant is shared across all token positions and all downstream components that read from it. Attention-head constants are fitted in the head's output space before $W_O$. We use Adam $\alpha = 0.002$ with batch of 16, each batch containing a single template and sequence length. We train for at least three epochs and stop when validation KL has not strictly decreased for three consecutive epochs. We report the test performance of the checkpoint with the lowest validation KL.

\begin{table*}[h]
\centering
\caption{Summary of attempts to improve exact reproduction of
model errors, with their correct-case effects and limitations.}
\label{tab:recovery_summary}
\begingroup
\small
\setlength{\tabcolsep}{5pt}
\renewcommand{\arraystretch}{1.15}
\begin{tabular}{@{}p{0.21\textwidth}p{0.35\textwidth}p{0.37\textwidth}@{}}
\toprule
Approach & Observed result & Tradeoff or limitation \\
\midrule

Full-training discovery
& Correct--error agreement gaps persist on Gemma ARC-Challenge
despite errors comprising 38.4\% of discovery examples.
& Greater discovery coverage has mixed effects across settings;
exposure to errors does not ensure their reproduction. \\
\addlinespace

Error reweighting
& IOI ACDC: assigning errors 50\% of discovery KL weight
raises $A_{\mathrm{err}}$ from 9.3\% to 37.0\%.
& $A_{\mathrm{ok}}$ falls from 98.7\% to 91.0\%. \\
\addlinespace

Distinct-error enrichment
& IOI ACDC: $A_{\mathrm{err}}$ rises from 9.3\% to 44.2\%
with 500 error and 500 correct discovery examples.
& $A_{\mathrm{ok}}$ falls from 98.7\% to 91.4\%;
enrichment changes discovery composition as well as
error prevalence. \\
\addlinespace

Optimal ablation
& Improves both agreement measures for some Docstring
circuits; large IOI agreement gaps remain.
& Benefits depend on the circuit and task; minimizing KL
does not ensure high exact-error agreement. \\
\addlinespace

Error-conditioned replacements
& Mean and resample replacements increase IOI
exact-error agreement.
& Scalar subject-logit shifts match or exceed recovery
at approximately matched correct-case cost.
These controls are calibrated on evaluation correct cases. \\
\addlinespace

Scalar and threshold calibration
& Can substantially increase IOI exact-error agreement;
matching overall error rates does not consistently
recover particular errors.
& Aggressive thresholds can narrow the agreement gap
by reducing correct preservation. \\
\addlinespace

Candidate-specific logit calibration
& Qwen ARC-Challenge: a fitted candidate bias raises
$A_{\mathrm{err}}$ from 38.5\% to 62.8\% and
$A_{\mathrm{ok}}$ from 59.3\% to 70.3\%
on 169 reserved validation questions.
& A successful partial recovery at the nominal 2\% edge
budget; an added output correction does not establish
the original circuit's causal fidelity. \\
\addlinespace

Targeted circuit expansion
& Adding 63 IOI head--position rules raises
$A_{\mathrm{err}}$ from 14.2\% to 75.1\% on the original
held-out evaluation.
& $A_{\mathrm{ok}}$ falls by 0.41 percentage points there
and 0.85 points on a fresh 5,200-prompt check, exceeding
the original 0.5-point tolerance on the fresh sample. \\

\bottomrule
\end{tabular}

\par\smallskip
\endgroup
\end{table*}

\paragraph{Attempts to recover model errors.}
Table~\ref{tab:recovery_summary} summarizes interventions targeting discovery data, ablation values, output predictions, and circuit membership. We test whether exposing extraction to more errors, increasing their discovery weight, changing replacement activations, calibrating circuit outputs, or restoring omitted components improves reproduction of the model's errors. Several interventions do produce substantial gains, but their correct agreement effects vary.

\section{MIB Data, Extraction, and Metric Implementation}
\label{app:mib_details}

\begin{table}[htbp]
\centering\small
\caption{Complete discovery coverage and candidate-scored public-test strata.}
\label{tab:appendix-mib-counts}
\begin{adjustbox}{max width=\textwidth}
\begin{tabular}{lrrrrrr}
\toprule
Setting & Train $N$ & Train correct & Train errors & Test $N$ & Test correct & Test errors \\
\midrule
Llama ARC-Challenge & 1119 & 908 & 211 & 586 & 464 & 122 \\
Gemma ARC-Challenge & 1119 & 689 & 430 & 586 & 349 & 237 \\
Qwen ARC-Challenge & 1119 & 659 & 460 & 586 & 315 & 271 \\
Llama ARC-Easy & 2251 & 2064 & 187 & 1188 & 1097 & 91 \\
Gemma ARC-Easy & 2251 & 1763 & 488 & 1188 & 927 & 261 \\
Llama subtraction & 17424 & 16575 & 849 & 1000 & 944 & 56 \\
\bottomrule
\end{tabular}
\end{adjustbox}
\end{table}

\paragraph{Dataset.} We use MIB's public prompts, splits and counterfactual fields \citep{mueller2025mib}.The evaluated curves come from the full public-training set. We report the full correct / error statistics in Table \ref{tab:appendix-mib-counts}.

\paragraph{Answer candidate and tie breaking.} ARC predictions select based on the highest logit among the prompt's answer-label tokens. Llama subtraction uses the single-token integer canddiates from 0 to 99. The original candidate order is the tie breaker: choice order for ARC and ascending numerical order for arithmetic. \textbf{This differs from the lowest token ID rule in the other tasks.}

\paragraph{Counterfactuals and circuit extraction.} We extract MIB circuits using EAP-IG \citep{hanna2024have} , with five input-interpolation steps and attribution scores computed over the entire public training split. For ARC counterfactual prompts replace the letter labels with numbers while preserving the question and answer choices. For subtraction, they replace the operands with randomly sampled values. We use these paired prompts for circuit scoring and evaluation, replacing excluded edge contributions with activations from the corresponding counterfactual run. 

\paragraph{Circuit budgets and MIB metrics.} Following MIB, we evaluate circuits at edge budgets $$s \in \{0.001, 0.002, 0.005, 0.01, 0.02, 0.05,0.1, 0.2, 0.5, 1\}$$. 
Let $F_j$ denote normalized task-score faithfulness at budget $s_j$.
We compute
\[
\mathrm{CPR}
= \sum_{j=1}^{9}(s_{j+1}-s_j)\frac{F_j+F_{j+1}}{2},
\qquad
\mathrm{CMD}
= \sum_{j=1}^{9}(s_{j+1}-s_j)
  \frac{|1-F_j|+|1-F_{j+1}|}{2}.
\]

\begin{table}[t]
\centering
\caption{
Public-test MIB aggregate task-score metrics
}
\small
\setlength{\tabcolsep}{5pt}
\begin{tabular}{llrr}
\toprule
Model & Task & CPR $\uparrow$ & CMD $\downarrow$ \\
\midrule
Llama-3.1 8B & ARC-Challenge & 0.8388 & 0.1602 \\
Gemma-2 2B   & ARC-Challenge & 1.0017 & 0.0547 \\
Qwen-2.5 0.5B & ARC-Challenge & 0.9364 & 0.0646 \\
Llama-3.1 8B & ARC-Easy      & 0.8458 & 0.1639 \\
Gemma-2 2B   & ARC-Easy      & 0.9943 & 0.0392 \\
Llama-3.1 8B & Subtraction   & 0.9954 & 0.0036 \\
\bottomrule
\end{tabular}
\label{tab:mib-aggregate}
\end{table}
Table ~\ref{tab:mib-aggregate} reports CPR (higher is better) and CMD (lower is better). 

\section{Margin-matched agreement}
\label{app:margin_matching}

\begin{table*}[t]
\centering
\caption{Agreement with $\mathcal{M}$ after matching correct
and errors model predictions on absolute answer margin.
Predictions are selected by argmax over the benchmark candidate set $\mathcal{K}(x)$;
Circuits and selection procedures are the same to Table~\ref{tab:main}.}
\label{tab:margin-matched}
\begingroup
\small
\setlength{\tabcolsep}{5pt}
\renewcommand{\arraystretch}{1.08}
\begin{adjustbox}{max width=\textwidth}
\begin{tabular}{@{}llrrr@{}}
\toprule
Circuit & Ablation &
\multicolumn{2}{c}{Matched agreement (\%)} &
$100\Delta_{\mathrm{match}}$ (pp) \\
\cmidrule(lr){3-4}
& & $A_{\mathrm{ok}}$ & $A_{\mathrm{err}}$ & [95\% CI] \\
\midrule

\multicolumn{5}{@{}l}{\textit{IOI}} \\
\addlinespace[2pt]

Manual reference & Mean
& 94.9 & 14.8 & $80.1\;[77.3,83.0]$ \\
& Resample (ABC)
& 71.3 & 40.0 & $31.3\;[26.7,36.0]$ \\
& OA (KL)$^\dagger$
& 98.4 & 10.2 & $88.2\;[86.0,90.4]$ \\
\addlinespace[2pt]

ACDC ($\tau=0.003$) & Mean
& 91.1 & 11.3 & $79.8\;[76.8,82.7]$ \\
& Resample (ABC)
& 80.9 & 24.1 & $56.8\;[52.7,60.7]$ \\
& OA (KL)$^\dagger$
& 96.3 & 8.3 & $88.0\;[85.7,90.2]$ \\
\addlinespace[2pt]

EAP-IG-KL & Mean
& 81.9 & 42.4 & $39.5\;[35.2,43.9]$ \\
& Resample (ABC)
& 73.4 & 51.8 & $21.6\;[17.0,26.1]$ \\
& OA (KL)$^\dagger$
& 92.3 & 45.8 & $46.5\;[42.7,50.2]$ \\
\addlinespace[2pt]

Edge Pruning & Mean
& 93.0 & 11.1 & $81.9\;[79.7,83.9]$ \\
& Resample (ABC)
& 84.3 & 17.3 & $67.0\;[63.8,70.1]$ \\
& OA (KL)$^\dagger$
& 92.7 & 15.4 & $77.3\;[75.0,79.6]$ \\

\midrule
\multicolumn{5}{@{}l}{\textit{Docstring}} \\
\addlinespace[2pt]

Manual reference & Mean
& 72.9 & 35.8 & $37.1\;[35.0,39.2]$ \\
& Resample
& 43.4 & 38.9 & $4.4\;[2.1,6.7]$ \\
& OA (KL)
& 77.3 & 42.4 & $34.8\;[32.8,36.9]$ \\
\addlinespace[2pt]

ACDC ($\tau=0.005$) & Mean
& 61.6 & 53.3 & $8.3\;[6.0,10.5]$ \\
& Resample
& 65.4 & 56.3 & $9.1\;[6.9,11.3]$ \\
& OA (KL)
& 79.6 & 41.5 & $38.2\;[36.1,40.2]$ \\
\addlinespace[2pt]

ACDC ($\tau=0.02$) & Mean
& 71.1 & 37.2 & $33.9\;[31.7,36.1]$ \\
& Resample
& 54.3 & 41.9 & $12.4\;[10.1,14.7]$ \\
& OA (KL)
& 73.9 & 39.0 & $34.9\;[32.7,37.0]$ \\
\addlinespace[2pt]

EAP-IG-KL & Mean
& 56.7 & 69.4 & $-12.6\;[-14.8,-10.5]$ \\
& Resample
& 57.3 & 66.3 & $-9.0\;[-11.1,-6.8]$ \\
& OA (KL)
& 85.9 & 60.0 & $25.9\;[24.1,27.7]$ \\

\bottomrule
\end{tabular}
\end{adjustbox}

\par\smallskip
\begin{minipage}{\textwidth}
\footnotesize
\end{minipage}
\endgroup
\end{table*}

\begin{figure}[t!]
\centering
\includegraphics[width=0.9\textwidth]{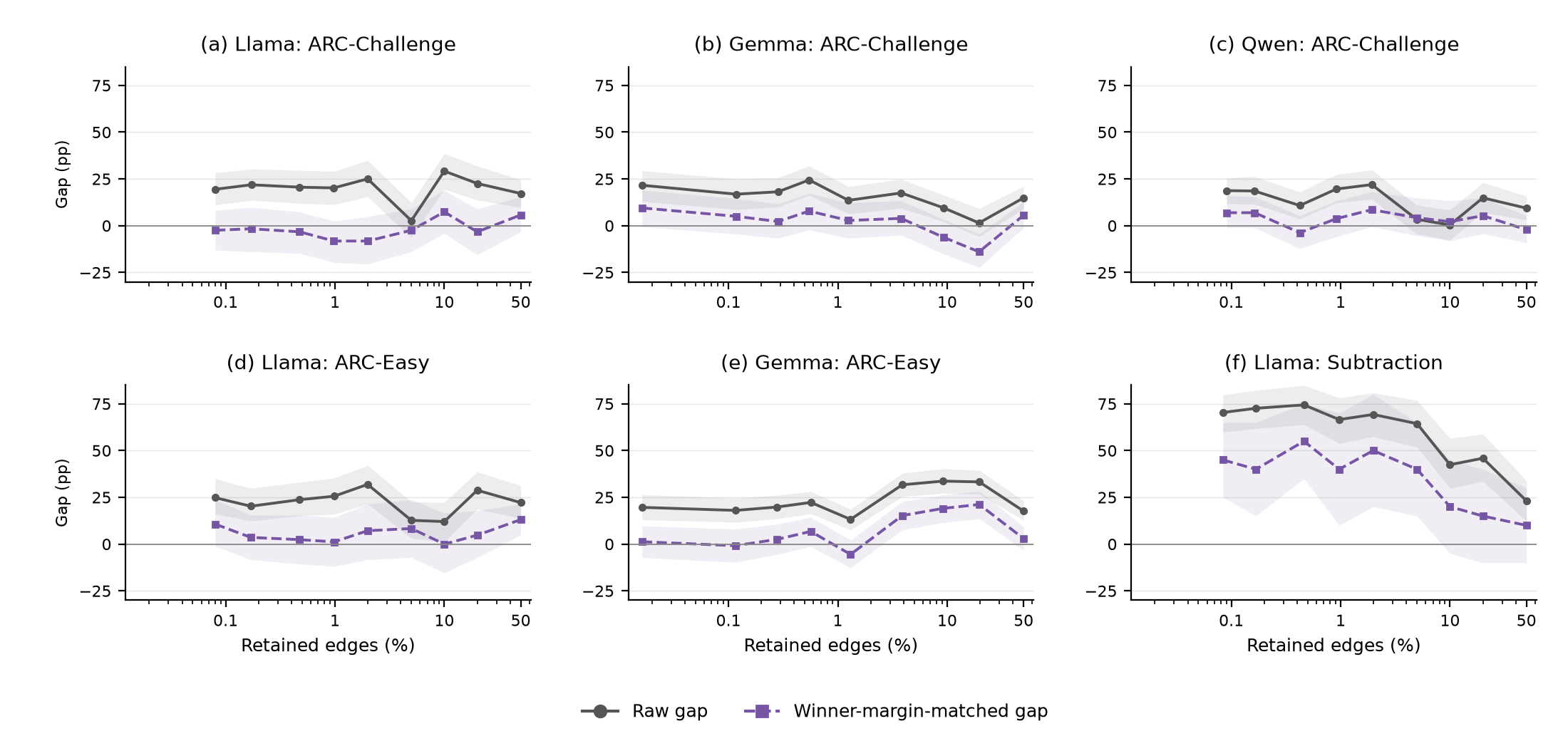}
\caption{Correct minus error agreement gaps on \textsc{MIB} before and after margin matching. Circuits are extracted using EAP-IG on the full public-training split and evaluated on the public-test split. Correct and error examples are bucketed within each model--task setting within a 0.1 logit range without replacement; Shading denotes pointwise 95\% bootstrap intervals.}
\label{fig:gap-margin-matched}
\end{figure}

We examine whether differences in intact-model prediction margins account for the gap between correct-case preservation and error reproduction. This analysis reuses the same evaluated circuits and their predictions. For all evaluations, predictions are selected through argmax over the benchmark candidate set $\mathcal{K}(x)$.

\paragraph{Matching procedure.}
For IOI and Docstring, we match on the highest-versus-second-highest candidate logits. Within each seetting, we pair model-error examples with model correct examples without replacement, within a $0.1$ logit slack. 

\paragraph{IOI and Docstring.} We use the absolute difference between the higehst and second-highest full model logits over $K(x)$. Matching still retains 805/884 IOI errors and 3,534/3,536 Docstring errors. All twelve IOI configurations still have positive matched gaps. Docstring EAP-IG-KL is an exception under mean and resample ablation, with negative percentage points. 

\paragraph{MIB.} Figure ~\ref{fig:gap-margin-matched} uses the top1 minus top2 candidate margin, with fixed pairs across size. Matching reduces several ARC gaps, but results still depend on setting and size. Subtraction margin matching retains only 20/56 errors. 

\section{IOI case-study details}

\begin{table}[h]
\centering
\small
\caption{IOI Manual Reference Circuit Heads}
\begin{tabular}{lp{0.53\linewidth}l}
\toprule
Reference group & Heads & Retained position \\
\midrule
Name Movers & $9.9,10.0,9.6$ & \textsc{END} \\
Backup Name Movers & $10.10,10.6,10.2,10.1,11.2,9.7,9.0,11.9$ & \textsc{END} \\
Negative Name Movers & $10.7,11.10$ & \textsc{END} \\
S-Inhibition & $7.3,7.9,8.6,8.10$ & \textsc{END} \\
Induction & $5.5,5.8,5.9,6.9$ & \textsc{S2} \\
Duplicate Token & $0.1,0.10,3.0$ & \textsc{S2} \\
Previous Token & $2.2,4.11$ & \textsc{S1+1} \\
\bottomrule
\end{tabular}
\label{tab:ioi_reference}
\end{table}

\subsection{Task, prediction rule, and reference circuit}
\label{app:ioi_terminology}
\paragraph{Task and terminology.} Indirect object identification (IOI) models the task of predicting the recipient of an action. For instance, in "Then, Sara and Courtney went to the hospital. Courtney gave a bone to", in this case, the correct indirect object (IO) is Sara, and the subject (S) is Courtney. We define the subject's first and second occurrences by S1 and S2, the position right after its first occurrence by S1+1, the final input position by END. For ABBA prompts, the first mentioned name is IO and the second mentioned name is S. For BABA prompts, the first two roles are reversed. The repeated name is S in both cases.

\paragraph{Exact answers and conditional agreement.} For every prediction in this case study, it is selected by argmax over the fixed, prompt-specific candidates [IO, S], with ties assigned to IO. Define
\begin{equation}
 m_C(x)=z_C(\mathrm{IO}\mid x)-z_C(\mathrm{S}\mid x),
 \qquad
 \widehat y_C(x)=
 \begin{cases}
 \mathrm{IO}, & m_C(x)\geq 0,\\
 \mathrm{S}, & m_C(x)<0.
 \end{cases}
\end{equation}

So, on model errors, a negative margin reproduces that particular wrong subject name. 

\paragraph{Reference implementation.} We use GPT-2 small in float32 and the published IOI head groups from \citet{wang2023interpretability}, instantiated as the 26 nodes in Table ~\ref{tab:ioi_reference}.

\paragraph{Data and search space.} We use a 60,000 prompt base, split by unordered name pair into 42,584 training prompts (463 model errors) and 17,416 validation prompts (177 errors).  The candidate space contains 3,574 supported additions across all 144 heads and 25 position rules: END, IO, S1, S2, S1+1, BOS, and offsets from END excluding the officially listed positions. Our term $P_{added}$ is the mean number of instantiated additional head outputs per prompt. 

\paragraph{Screening candidate nodes.}
Evaluating every possibel addition at every search step is expensive. Therefore, we use gradients to approximate promising additoins, then we evaluate the candidates exactly. 
We represent the circuit with gates $\mathbf g,$ one for each head--position:
\begin{equation}
\widetilde{\mathbf o}_i(x;\mathbf g)
=
\boldsymbol\mu_{i,t(x)}
+
g_i\bigl(\mathbf o_i(x;\mathbf g)-\boldsymbol\mu_{i,t(x)}\bigr),
\end{equation}
where $\mathbf o_i$ is the output of the intervened circuit and
$\boldsymbol\mu_{i,t(x)}$ is its template-matched ablation mean.

At each search step, we temporarily relax the candidate gates to $[0,1]$
and differentiate the full-training loss
\begin{equation}
\mathcal L(\mathbf g)
=
\underset{x\in\mathcal E_{\mathrm{tr}}}{\operatorname{mean}}
\operatorname{softplus}\bigl(m_{C(\mathbf g)}(x)\bigr)
+
w\underset{x\in\mathcal O_{\mathrm{tr}}}{\operatorname{mean}}
\operatorname{softplus}\bigl(-m_{C(\mathbf g)}(x)\bigr),
\end{equation}
where $\mathcal E_{\mathrm{tr}}$ and $\mathcal O_{\mathrm{tr}}$ are the
original model errors and correct examples, respectively.
We set $w=8$ when correct agreement falls more than $0.005$ below
the reference circuit, and $w=1$ otherwise.

We rank omitted outputs by
\begin{equation}
s_i
=
-\left.
\frac{\partial\mathcal L(\mathbf g)}{\partial g_i}
\right|_{\mathbf g=\mathbf g_{\mathrm{current}}},
\end{equation}
These proposals are then evaluated on the complete training prompt set with binary gates.

\paragraph{Evaluating and selection additions.}
At each search step, we test each of the four highest-scoring additions
individually, every pair among those four, and three larger proposals
containing the top 4, 8, or 16 additions. We remove duplicate proposals
and evaluate each remaining circuit on the entire training set, with
every gate set to either zero or one. We select the proposal by maximizing:
\begin{equation}
J(C)
=
A_{\mathrm{err}}(C)
-8\bigl[A_{\mathrm{ok}}(C_0)-A_{\mathrm{ok}}(C)-0.005\bigr]_+
-\lambda_P P_{\mathrm{added}}(C).
\end{equation}
This score rewards exact error reproduction while penalizing correct agreement drops, and penalizes big extension sizes. We run two searches from $C_0$, using
$\lambda_P\in\{0.0005,0.002\}$, each with at most 32 addition rounds
and 128 added rules. 

During each search, we also test removing previous  added nodes, either individually or together with other additions part of the same head. We accept a removal only if it maintains or improves $J(C)$, while keeping the original reference circuit intact.

\paragraph{Choosing the final extension.}
On validation, we keep candidates whose correct agreement is at most $0.005$ below $C_0$. Among these candidates, let $G^\star$ be the largest improvement in exact error agreement over $C_0$. We select candidates that minimize $P_{\mathrm{added}}$ subject to retaining at least $90\%$ of $G^\star$ and breaking ties by higher error agreement.

Then, we evaluate every added component and every joint delete of a head's added rules. After adding the candidates, we recompute $G^\star$, and reapply the same selection process until no tested deletion yields a smaller circuit satisfying both agreement thresholds. Convergence happened after three sweeps to 63 added rules. 

Finally, we freeze this extension and evaluate it on a 26,000 hold out set with 225 model errors and 25,775 model correct examples.

\end{document}